\documentclass{article}

 \usepackage[preprint]{neurips_2026}

\usepackage[utf8]{inputenc} 
\usepackage[T1]{fontenc}    
\usepackage{hyperref}       
\usepackage{url}            
\usepackage{booktabs}       
\usepackage{amsfonts}       
\usepackage{nicefrac}       
\usepackage{microtype}      
\usepackage{xcolor}         
\usepackage{amsmath}        
\usepackage{lineno}
\usepackage{multirow}
\usepackage[table]{xcolor}
\usepackage{graphicx}
\usepackage{tcolorbox}

\definecolor{darkblue}{rgb}{0, 0, 0.5}

\title{Boundary-targeted Membership Inference Attacks on Safety Classifiers}

\author{%
  Anthony Hughes\thanks{Corresponding author: \texttt{ajhughes3@sheffield.ac.uk}} \\
  University of Sheffield\\
  \texttt{ajhughes3@sheffield.ac.uk} \\
  \And
 Alexander Goldberg \\
 Carnegie Mellon University \\
 \texttt{akgoldbe@andrew.cmu.edu} \\
  \And
  Prince Jha \\
  MBZUAI \\
  \texttt{prince.jha@mbzuai.ac.ae} \\
  \And
  Adam Perer \\
  Carnegie Mellon University \\  
  \texttt{adamperer@cmu.edu} \\
  \And
  Nikolaos Aletras \\
  University of Sheffield \\
  \texttt{n.aletras@sheffield.ac.uk} \\
  \And
    Niloofar Mireshghallah \\
  Carnegie Mellon University \\
  \texttt{nmireshg@andrew.cmu.edu} \\
}

\begin{document}

\maketitle

\begin{abstract}
Safety classifiers are essential safeguards within generative AI systems, filtering harmful content or identifying at-risk users when interacting with large language models.
Despite their necessity, these models are trained on sensitive datasets including discussions of self-harm and mental health, raising important, yet poorly understood, privacy concerns.
Membership inference attacks (MIAs) allow adversaries to infer membership of examples used to train models.
In this work, we hypothesize that identifying the examples on which the classifier is least confident are informative for an adversary to infer membership. 
This reflects a localized failure of generalization, where the model relies on memorization to resolve ambiguity in the training set.
To investigate this, we introduce a new boundary-targeted selection strategy that identifies low confidence examples that amplify the signal of an examples membership within a training set. 
Our experimental results show that an adversary can recover 19\% of the conversations a safety classifier flagged as indicating user distress, at a 5\% false-positive rate, on a classifier fine-tuned for detecting a user who may require emotional support.
This is $3.5$ times more than attacking using state-of-the-art MIA methods alone.
Finally, we characterize the boundary laying examples and show that content-based filtering is ineffective for protection, and existing noise strategies can effectively mitigate susceptibility of these examples.\footnote{Our code is available at \url{https://github.com/anthonyhughes/safety-classifiers}}
\end{abstract}

\section{Introduction}
\label{intro}
Large language models (LLMs) have demonstrated remarkable capabilities across a wide range of tasks \citep{dubey_llama_2024, team_olmo_2025}, yet their deployment in user-facing applications raises safety concerns \citep{dong_position_2024, gangavarapu_enhancing_2024, hakim_need_2024, naseem_gametox_2025}.
These concerns range from the generation of toxic or harmful content\citep{chao_jailbreakbench_2024, mazeika_harmbench_2024, mehrotra_tree_2024, naseem_gametox_2025} to the amplification of distress in users disclosing personal vulnerabilities \citep{zhao_wildchat_2024,mireshghallah_trust_2024}. 
To identify at-risk behavior during interaction LLM providers increasingly rely on safety classifiers to intercept and flag harmful inputs before they reach the underlying foundational model \citep{cunningham_cost-effective_2025,sharma_constitutional_2025, kramar_building_2026}.
While effective at content moderation, these classifiers are increasingly being adapted in more user welfare-oriented settings \citep{farinhas_mindguard_2026}.  Detecting crises or delusional episodes requires authentic examples of such phenomena that originates from vulnerable individuals \citep{cohan_smhd_2018, mireshghallah_can_2024, siddals_it_2024, morrin_artificial_2026}, and therefore carries high re-identification risk \citep{lermen_large-scale_2026, li_agentic_2026}.

Practically, safety models must satisfy several requirements reflecting the diversity of harms that arise in real user interactions \citep{wang_self-guard_2024, lv_gamma-guard_2025, upadhayay_x-guard_2025}. 
At the level of a single-turn conversation, they must detect toxic prompts that seek to produce content that facilitates harassment, blackmail, or violence \citep{inan_llama_2023,openai_technical_2025}. 
Increasingly, these systems are also expected to operate at a longitudinal scale \citep{kramar_building_2026}, identifying patterns of sustained distress, emotional dependency, or deteriorating well-being across sessions \citep{cheng_sycophantic_2025, kwik_digital_2025}, all using training data drawn from real users. Despite this fact, there is little work examining whether safety classifiers leak information about these individuals. In this paper we present the first systematic study of membership inference attacks (MIA) against safety classifiers, characterizing which training examples are exposed and proposing defenses.

MIAs are statistical methods for determining whether specific text examples (i.e., documents) and private attributes (i.e., labels) were included in a model's training data \citep{shokri_membership_2017,feldman_does_2020, song_overlearning_2020, choquette-choo_label-only_2021, carlini_membership_2022, mireshghallah_quantifying_2022, chang_context-aware_2025, chen_statistical_2025, hallinan_surprising_2025}. 
Beyond the risk of verbatim data extraction \citep{carlini_extracting_2021, lukas_analyzing_2023}, a critical concern is whether the properties that render safety classifiers effective may simultaneously amplify their susceptibility to membership inference. Conventionally, MIA techniques target high-confidence examples, those the classifier handles most effectively, as the primary candidates for revealing memorization \citep{carlini_membership_2022}.

Contrary to this convention, we hypothesize that for safety classifiers informative targets for membership inference lie at the opposite end of the confidence distribution. We argue that ambiguous examples (i.e., on which the classifier is naturally least confident) are the instances where high-confidence predictions are most clearly attributable to memorization rather than generalization. Safety classification is inherently characterized by contested cases where the boundary between safe and unsafe content is not easily distinguishable, annotator agreement is low, and harm is often subtle \citep{leonardelli_agreeing_2021, fleisig_when_2023, wan_noise_2025}. 

Building on this insight, we introduce \textbf{\emph{boundary-targeted selection}}: a MIA strategy that isolates membership evaluation to examples residing near the classifier's decision boundary, where the membership signal is concentrated. We evaluate this strategy across two MIA scoring functions, four language models, and five datasets spanning toxicity detection, jailbreak resistance, and mental-health classification, finding consistent and substantial improvements in attack effectiveness over standard example selection methods. We additionally characterize the textual and semantic properties of boundary-vulnerable examples, showing that content-based filtering is ineffective as a defense, and propose an inference-time noise-injection strategy that mitigates the attack.

\section{Related Work}

\paragraph{Privacy and Inference Attacks.} 
LLMs are vulnerable to privacy attacks \citep{carlini_extracting_2021, yu_bag_2023, nasr_scalable_2025, hayes_measuring_2025}. Prior work demonstrates that training data can be extracted directly from model outputs \citep{nasr_scalable_2025}. This vulnerability is largely attributable to memorization, i.e. models disproportionately retain and repeat seen training examples \citep{carlini_quantifying_2022, feldman_does_2020, zhang_membership_2025}. Beyond verbatim extraction, recent work has shown that training data can be reconstructed \citep{elmahdy_deconstructing_2024}, and that models leak sensitive attributes about individuals even without recovering exact text \citep{staab_beyond_2024}. Inference attacks are methods that determine whether a specific document or its label was present in training data \citep{shokri_membership_2017, song_overlearning_2020, carlini_membership_2022,  chaudhari_snap_2023, puerto_scaling_2025,rossi_membership_2025, marek_benchmarking_2026}. While these attacks have been applied to general-purpose classifiers settings \citep{shejwalkar_membership_2021, mireshghallah_quantifying_2022, xie_recall_2024} their implications for safety specific classifiers remain unexplored.

\paragraph{Safety Classifiers.} Safety classifiers are discriminative models trained on user interaction data. These are deployed by model providers to intercept policy-violating inputs at inference time \citep{inan_llama_2023, openai_technical_2025, zhao_qwen3guard_2025}. When a classifier labels an input as unsafe, the platform typically refuses to continue the conversation. These classifiers are first line of defense for providers \citep{sharma_constitutional_2025, kramar_building_2026}, typically categorizing inputs of harm like harmful, offensive, or illegal content. Prior work has focused on single and multi-turn jailbreaks, where adversarial users construct conversational contexts designed to elicit policy-violating outputs \citep{perez_red_2022, wei_jailbroken_2023, cunningham_cost-effective_2025}. An emerging class of safety concerns goes beyond single conversation interactions, users increasingly turn to LM-based platforms for emotional support, disclosing mental health struggles, suicidal ideation, and longitudinal distress across sessions \citep{liu_towards_2021, siddals_it_2024, mireshghallah_trust_2024}. Detecting and appropriately responding to these signals requires classifiers that operate over extended interaction histories rather than individual turns \citep{morrin_artificial_2026}. Welfare monitoring classifiers will require labeled examples of crisis disclosures likely tied to real individuals. In this paper, we look to study how these scenarios may lead to exposure via MIAs.

\section{Problem Formulation}
\label{sec:prelims}

\begin{figure}
    \centering
    \includegraphics[width=1\linewidth]{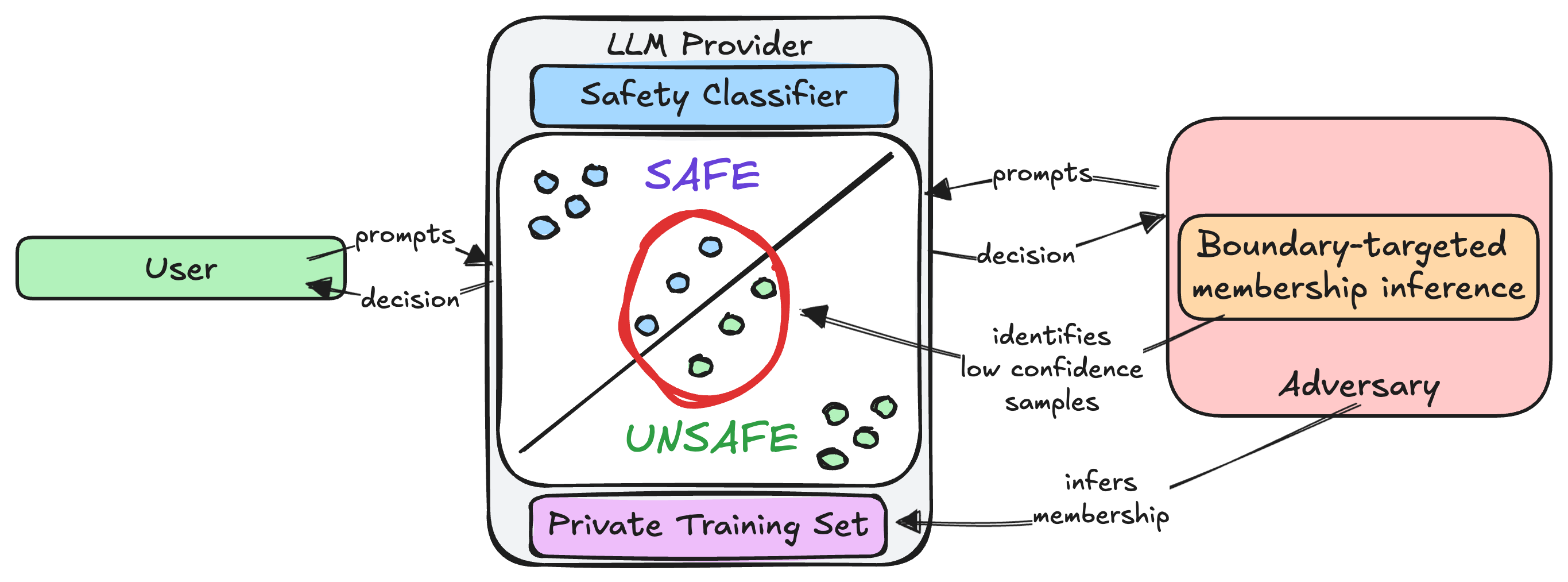}
    \caption{
    Overview of the threat model.
    An LLM provider deploys a safety classifier $\mathcal{C}$, trained on a private corpus $\mathcal{D}$ of labeled user conversations, to flag inputs as \textsc{safe} or \textsc{unsafe}.
    An adversary, acting as a regular user, submits prompts and observes the classifier's decisions.
    Using this information the adversary mounts a boundary-targeted membership inference attack.
    \emph{Our key finding is that knowledge of where the decision boundary lies relative to training-set members enables the attacker to infer whether specific conversations were present in $\mathcal{D}$.}
    }
    \label{fig:attack-overview}
\end{figure}

We assume an LLM provider deploys a binary safety classifier over user conversations. For example, a mental-health risk classifier predicting whether a dialogue indicates risk of self-harm. The classifier $f_\theta$ is trained on a private dataset $D = \{(x_i, y_i)\}_{i=1}^n$ of past user conversations, where each $x_i$ is a conversational input and each $y_i \in \mathcal{S}$ is a safety label drawn from a binary label set $\mathcal{S} = \{\text{safe}, \text{unsafe}\}$.

An external adversary queries the deployed classifier with candidate conversations and observes the predicted class-probability vector $f_\theta(x) \in [0,1]^{|\mathcal{S}|}$, from which the predicted label $\hat{y}(x) = \arg\max_{s \in \mathcal{S}} f_\theta(x)_s$ can be recovered. The adversary's goal is to determine, given a candidate labeled example $(x, y)$, whether it was a member of the training set $D$. The same threat model applies to other safety classifiers trained on sensitive conversational data, including emotional-support and toxicity classifiers. \autoref{fig:attack-overview} illustrates one concrete instantiation of the threat model studied in this paper.

Prior work has shown that access to a reference or shadow classifier improves attack effectiveness \citep{shokri_membership_2017, carlini_membership_2022}. We therefore equip the adversary with a reference classifier $f_{\mathrm{ref}}$, trained on data drawn from a distribution related to that of $D$. This assumption is realistic in our setting: model providers routinely release public safety-classifier weights, including Llama Guard \citep{inan_llama_2023} and ShieldGemma \citep{zeng_shieldgemma_2024, zeng_shieldgemma_2025}, which an adversary can use directly as a reference without access to the target's training data.

\section{Boundary-targeted Membership Inference}
\label{sec:mia}

Our central methodological contribution is \emph{boundary-targeted selection}, an example selection strategy that filters the evaluation pool to examples near the classifier's decision boundary. Standard membership inference attacks evaluate over uniformly sampled examples, on the assumption that the strongest membership signal lies in examples the classifier handles unusually well. We hypothesize that for safety classifiers, the more informative targets lie at the opposite end of the confidence distribution. This hypothesis is grounded in how state-of-the-art attacks operate. Likelihood-ratio attacks (LiRA, \citealp{carlini_membership_2022}) calibrate the target's confidence against a reference model trained on disjoint data, which reduces signal on examples both classifiers find easy. The residual signal therefore lives at the boundary, where the reference has no independent reason to be confident and any remaining confidence in the target is most plausibly attributed to memorization rather than the example's intrinsic separability. To evaluate this strategy we instantiate two standard scoring functions from prior work, both operating under the score-based access of our threat model.

\subsection{Membership Inference Setup}

Following prior work \citep{shokri_membership_2017, carlini_membership_2022}, we partition the data into five disjoint splits with balanced safe and unsafe classes: training splits $D_A$ and $D_B$, a validation split $D_\text{val}$, a calibration split $D_\text{cal}$, and an evaluation split $D_\text{eval}$. Given an open-weight LLM, we use it to fine-tune two different binary classifiers, $f_A$ and $f_B$, with identical architecture and hyperparameters on $D_A$ and $D_B$ respectively. To attack $f_A$, we treat it as the target $f_\theta$, draw members from $D_A$ and non-members from $D_B$, and use $f_B$ as the reference $f_{\text{ref}}$. The symmetric procedure attacks $f_B$. Because $D_A \cap D_B = \emptyset$, the reference has no training signal on members of the target.

\subsection{Boundary-targeted Selection}

Operationally, the adversary retains global access to the full candidate pool of members and non-members but adds a filtering step. A selection classifier $f_S$ is trained on $D_\text{cal}$, a calibration set disjoint from both $D_A$ and $D_B$, making $f_S$ independent of the target and reference. The independence of $f_S$ from the target and reference is deliberate to avoid the circularity in which the same model defines both the boundary and the membership signal.  The adversary then scores every candidate by the probability $f_S$ assigned to the ground-truth label, $P_S(y_i \mid x_i) = f_S(x_i)_{y_i}$. For each label $y \in \mathcal{S}$ the adversary retains the $n$ candidates with the lowest $P_S(y_i \mid x_i)$, giving the boundary set $\mathcal{B}$. The parameter $n$ controls the size of $\mathcal{B}$ and can be tuned to trade off between signal strength and statistical power. Selecting equally across labels prevents any single class from dominating the boundary set. 

Both global and boundary-targeted selection are applied independently to members and non-members, and the scoring functions described below are evaluated on both selection strategies.

\subsection{Scoring Functions}

Each scoring function assigns a membership signal $s(x, y)$ to a candidate $(x, y)$, with larger values indicating greater evidence of membership. The two scoring methods differ only in the auxiliary resources required.\footnote{We also make results of a shadow-based attack \citep{shokri_membership_2017} in \autoref{app:results:uber}.}

\paragraph{Loss-based.} \cite{yeom_privacy_2018} thresholds the target classifier's confidence on the true label, $s_\text{loss}(x, y) = \log f_\theta(x)_y$, and is the baseline that mimics an adversary with minial resources.

\paragraph{LiRA.} \citep{carlini_membership_2022} calibrates the target's confidence against the reference, $s_\text{LiRA}(x, y) = \log f_\theta(x)_y - \log f_{\text{ref}}(x)_y$, controlling for example difficulty. We use $f_B$ as the reference when attacking $f_A$, and vice versa.

\subsection{Target Classifiers}

Deployed safety classifiers operate across a range of input structures. We study four target classifier configurations, each reflecting a distinct deployment scenario.

\paragraph{Single-turn classifier.} The simplest deployment scenario is a classifier that flags individual user prompts in isolation. Most safety filters operate at this level, evaluating each prompt or prompt-response pair before further processing \citep{ji_beavertails_2023}. An adversary inferring membership of a single prompt reveals that a specific user submitted that prompt and that the classifier assigned it a particular safety label.

\paragraph{Multi-turn classifier.} As conversational interactions extend, safety classifiers increasingly operate over multi-turn dialogues to capture context that single-turn evaluation misses. Multi-turn jailbreaks, for instance, can spread harmful intent across multiple exchanges that each appear benign in isolation \citep{rahman_x-teaming_2025}. Privacy risk here is not just an isolated prompt but a sequence of exchanges that together identify a user's interaction.

\paragraph{Multi-conversation classifier.} Frontier labs are increasingly deploying safety classifiers that operate at a longitudinal scale, identifying patterns of sustained distress, emotional dependency, or deteriorating well-being across a user's recent interaction trajectory \citep{kramar_building_2026, cheng_sycophantic_2025}. Membership inference reveals the longitudinal pattern of an individual's interaction with the system, including conversations the user did not intend any single classifier decision to depend on.

\paragraph{Pooled classifier.} Deployed safety classifiers may need to handle heterogeneous input structures within a single model rather than maintaining separate per-type classifiers \citep{cunningham_cost-effective_2025, cunningham_constitutional_2026, kramar_building_2026}. We study a pooled configuration that combines all three input structures within a single classifier to test whether attack signal transfers across input structures.

\section{Experimental Setup}

\subsection{Data}
\label{data:data}

We use publicly released datasets to instantiate each target classifier configuration described above.

\paragraph{Single-turn data.} \textbf{BeaverTails} \citep{ji_beavertails_2023} is a dataset of single-turn prompt-response pairs, each annotated with an overall safe/unsafe label as provided by the original dataset.

\paragraph{Multi-turn data.} \textbf{X-Guard-Train} \citep{rahman_x-teaming_2025} is a multi-turn dataset of jailbreak attempts and toxic dialogues. We pair the unsafe data with multi-turn safe data drawn from \textbf{WildChat} \citep{zhao_wildchat_2024}.

\paragraph{Multi-conversation data.} The unsafe class is drawn from two sources. \textbf{ESConv} \citep{liu_towards_2021} is a crowd-sourced corpus of emotional-support dialogues in which a help-seeker discloses an emotional problem to a trained supporter. We use bundled conversations from these help-seekers as examples of users in emotionally vulnerable states. \textbf{Psychotherapy Eval} \citep{steenstra_assessing_2026} consists of therapy session transcripts centered on adverse events requiring therapist intervention; we concatenate sessions per patient as examples of users in clinically vulnerable contexts. The safe class is drawn from \textbf{WildChat}, restricted to conversations flagged as safe and bundled by hashed user identifier. Each multi-conversation input is a temporally ordered concatenation of conversations from a single user. We ensure that no conversation appears in both the multi-turn and multi-conversation training sets. 

\paragraph{Pooled data.} The pooled classifier is trained on the union of the data above. To prevent leakage across settings, WildChat is split disjointly between its multi-turn and multi-conversation roles before bundles are constructed, so no underlying conversation appears in more than one bundle across the pooled training set.

Finally, for each example $(x_i, y_i)$ in the dataset the document $(x_i)$ is inserted into a chat template provided in \autoref{app:prompts}. Additionally, we provide data split information in \autoref{app:data:splits}, general corpus statistics of all datasets in \autoref{app:table:corpus-stats}, context-length requirements \autoref{app:table:seq_len}, and examples of each dataset in \autoref{app:data:examples}.

\subsection{Models}
\label{sec:models}

We use three instruction fine-tuned Llama variants of different sizes: \textbf{Llama-3.2-1B-Instruct}, \textbf{Llama-3.2-3B-Instruct} and \textbf{Llama-3.1-8B-Instruct} \citep{dubey_llama_2024}. To assess whether our method generalizes across model families, we additionally employ \textbf{Gemma-3-12B-Instruct} \citep{team_gemma_2025}.

\subsection{Training Configuration}
\label{sec:training-config}

All safety classifiers are fine-tuned with and without LoRA \citep{hu_lora_2019}, following the current state-of-the-art in user moderation \citep{sharma_constitutional_2025, cunningham_constitutional_2026}. We target the query and value projection matrices $\mathbf{W}_Q$ and $\mathbf{W}_V$, and apply the same configuration to the target classifiers $f_A$ and $f_B$, the selection classifier $f_S$. Split sizes, LoRA scaling parameters, epoch counts, and per-model-size configurations are reported in \autoref{app:tab:hyperparameters}.

\subsection{Evaluation}
\label{sec:evaluation}

In order to evaluate our boundary-targeted MIA attacks on safe- and unsafe-labeled training data we require metrics to measure attack efficacy. For each paired classifier, we construct member and non-member sets from the disjoint training splits. When attacking $f_A$, members are drawn from $D_A$ and non-members from $D_B$; when attacking $f_B$, members are drawn from $D_B$ and non-members from $D_A$. Per-dataset training-split sizes are reported in \autoref{app:data:splits}.

Following prior work \citep{wen_canary_2023, zhang_membership_2025} we measure attack effectiveness using two metrics. \textbf{MI-AUC} (Membership Inference Area Under the ROC Curve) summarizes overall attack effectiveness across all decision thresholds, with higher values indicating greater accuracy in identifying training-set membership. We report MI-AUC under both global and boundary-targeted selection. \textbf{TPR at 5\% FPR} measures the fraction of genuine training members the attack correctly identifies when 5\% of flagged candidates are false alarms. A TPR of 30\% at 5\% FPR, for instance, indicates that an adversary can correctly recover 30\% of training members from a candidate pool while keeping the false-alarm rate at or below 5\%. To quantify uncertainty in both metrics, we report 95\% bootstrap confidence intervals over 1{,}000 resamples.


\section{Results}
\label{sec:results:classifiers}

\begin{figure}[t]
    \centering
    \includegraphics[width=1\linewidth]{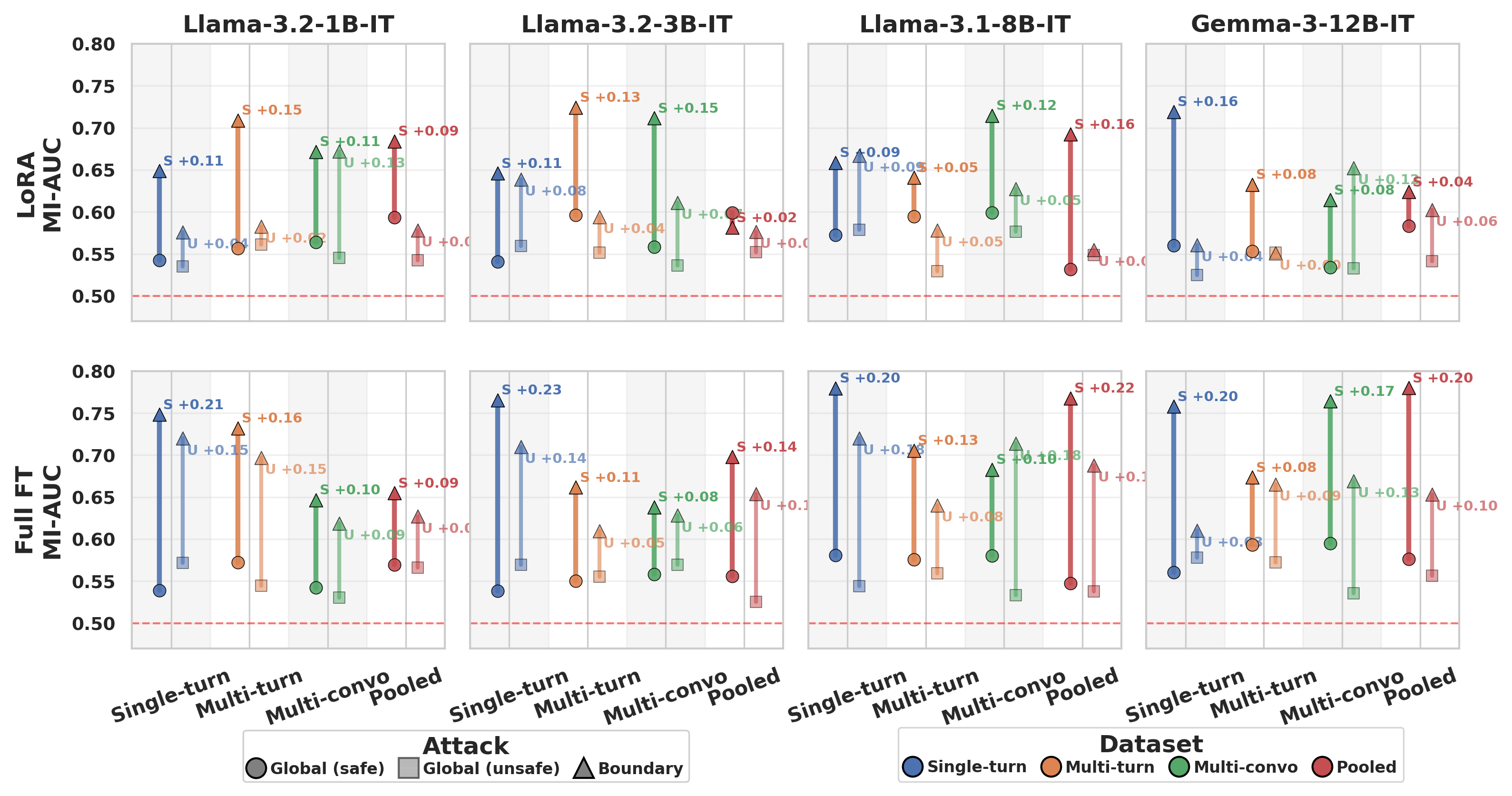}
    \caption{
    MIA performance on LiRA and boundary-targeted LiRA across model scales and training configurations.
    Boundary amplification is stratified by safe and unsafe labels.
    Dumbbells indicate the MI-AUC gain of boundary-targeted LiRA (triangles) over global LiRA (circles) for safe-labeled and unsafe-labeled samples.
    S and U annotations show the absolute AUC improvement for safe and unsafe training sample membership respectively.
    The horizontal dashed line denotes random-chance performance.
    }
\label{fig:safe_unsafe_boundaries}
\end{figure}

\autoref{fig:safe_unsafe_boundaries} shows the results of our boundary-targeted LiRA-based MIA attacks on safe- and unsafe-labeled training data.\footnote{A full set of results across all MIA selection strategies and scoring functions is provided in \autoref{app:results:uber}}.

\subsection{Effectiveness of Boundary-targeted MIAs}

\paragraph{Boundary-targeted samples are vulnerable.} Our boundary-targeted attack amplifies membership inference substantially. For the safe class, boundary thresholding yields gains of \textit{+0.12} to \textit{+0.16} MI-AUC across models under LoRA, and \textit{+0.02} to \textit{+0.13} under full fine-tuning. For multi-conversation data, which contains highly sensitive interactions, MI-AUC exceeds \textit{0.75} for unsafe conversations. We observe this increase in attack effectiveness across all models, training regimes, and datasets. Some configurations are more resistant than others. Llama-3.1-8B-Instruct, for example, is less susceptible to unsafe single-turn MIAs suggesting that model-specific properties also influence attack success.

\paragraph{Boundary knowledge as an exploit remains intact regardless of training capacity.} Current intuition is that parameter-efficient fine-tuning should reduce memorization by limiting model capacity \citep{liu_differentially_2025}. Our results support this for global LiRA, where LoRA-trained classifiers are less vulnerable than fully fine-tuned ones, particularly on unsafe content. However, the boundary-targeted attack closes this gap. For the safe class under LoRA, boundary gains of \textit{+0.12} to \textit{+0.16} match or exceed the gains observed under full fine-tuning (\textit{+0.02} to \textit{+0.13}). This suggests that LoRA's protective effect is concentrated at the population level, where easy examples remain safe, while the boundary signal that the targeted attack exploits remains intact.


\paragraph{LoRA conceals unsafe leakage rather than preventing it.} The unsafe class is where boundary targeting most clearly breaks the protective story around parameter-efficient fine-tuning. Under LoRA, global attacks struggle to recover unsafe membership at all, and the class appears largely resistant. Boundary targeting reverses this finding. It extracts \textit{+0.04} to \textit{+0.08} in additional MI-AUC under LoRA and \textit{+0.08} to \textit{+0.18} under full fine-tuning, identifying the same toxic and self-harm content that the global attack misses.

\begin{figure}[ht]
    \centering
    \includegraphics[width=1\linewidth]{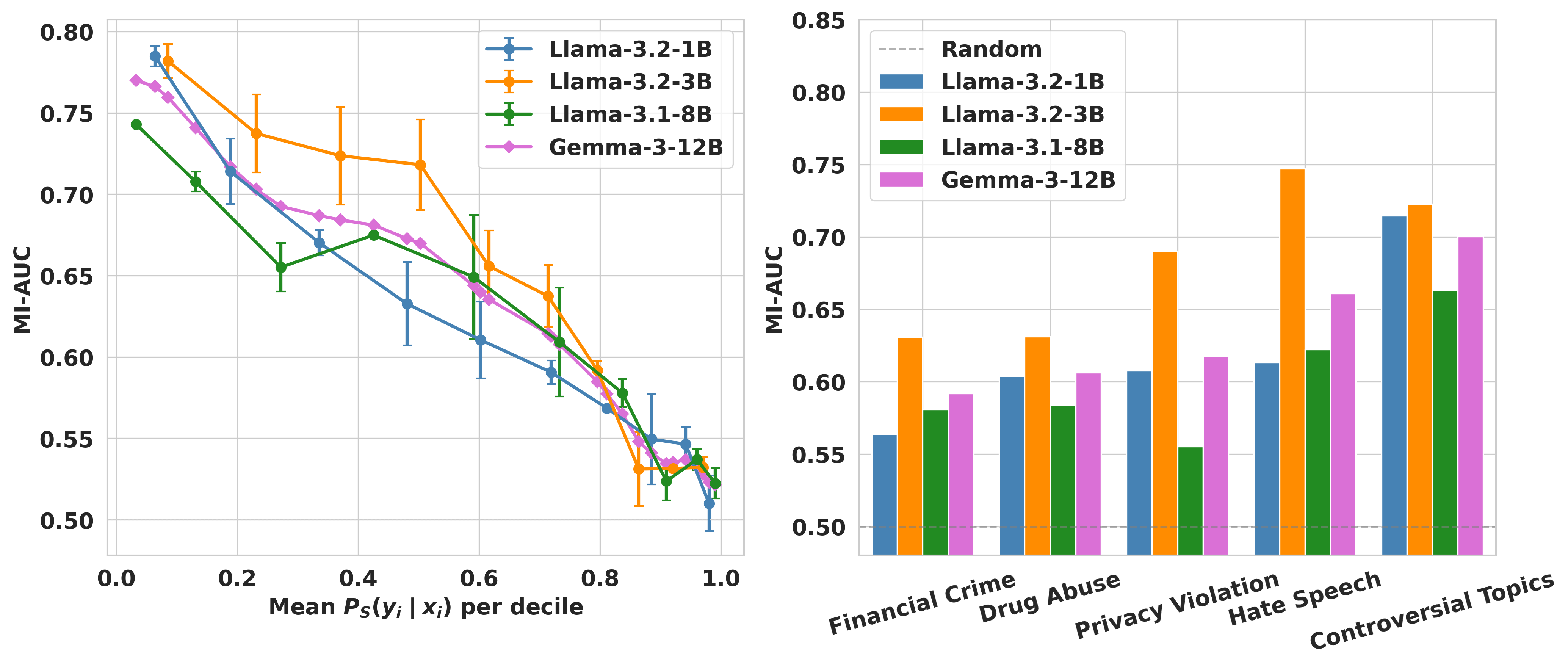}
    \caption{
    \textbf{(Left)} Boundary-targeted LiRA MI-AUC as a function of the classifier's true-label confidence $P_S(y_i \mid x_i)$, binned into deciles. 
    Each line corresponds to a model. Error bars span the min and max across the two classifiers.
    \textbf{(Right)} Boundary-targeted LiRA MI-AUC across the harm categories assigned to BeaverTails. 
    Bars show the mean MI-AUC averaged over all training regimes for each model. 
    \textbf{(Both)} A dashed grey line denotes random-chance performance.
    }
\label{fig:confidence_and_categories}
\end{figure}

\paragraph{High MI-AUC in low confidence samples across all models.} In \autoref{fig:confidence_and_categories} (left), we observe a consistent monotonic relationship between target-classifier confidence on the true label and membership inference effectiveness. Examples in the lowest confidence decile, those the classifier struggles with most, yield the highest MI-AUC across all four models, while examples in the highest decile, on which the classifier is most confident, sit closer to random chance. This pattern holds across every model we evaluate, indicating that boundary memorization is not specific to any model scale or architecture family. The shape of the curve is precisely what the boundary-targeted attack exploits. By filtering the evaluation pool to the lowest-confidence deciles, the adversary concentrates on the regime where the membership signal is strongest.

\paragraph{Vulnerability varies substantially across harm categories.} In \autoref{fig:confidence_and_categories} (right), we report MI-AUC by BeaverTails harm category for the three least and two most vulnerable categories across our models. `Controversial Topics' yields the strongest membership signal across all four models, with MI-AUC ranging from $0.66$ to $0.73$, and `Hate Speech' reaches a peak of $0.74$ under Llama-3.2-3B-Instruct. By contrast, `Financial Crime', `Drug Abuse', and `Privacy Violation sit closer to random chance for the same model where MI-AUC is between $0.55$ and $0.69$. This suggests that categories involving subjective judgments can leak more membership signal than categories with clearer harm definitions, consistent with our broader observation that ambiguous content can drive memorization.

\paragraph{Pooling datasets reduces but does not eliminate attacks.} The pooled classifier represents the realistic deployment scenario in which a single safety classifier is trained on heterogeneous input structures. Under LoRA, pooling reduces attack effectiveness modestly compared to the per-dataset configurations, with boundary-targeted gains of \textit{+0.04} to \textit{+0.16} on the safe class and \textit{+0.02} to \textit{+0.06} on the unsafe class. The reduction is most visible on multi-conversation content, where per-dataset boundary gains were substantially higher. Under full fine-tuning, however, pooling provides no such protection. Boundary-targeted gains reach \textit{+0.22} on the safe class and \textit{+0.15} on the unsafe class, and absolute MI-AUC approaches \textit{0.80} in several configurations.


\section{Protection Against Boundary-targeted MIA}

We find that boundary-targeted examples, those on which the classifier assigns the lowest confidence to the ground-truth label, are disproportionately vulnerable to MIA. In this section, we propose two mitigation strategies against this vulnerability. First, we hypothesize that boundary examples can be filtered from the training set, motivated by works distinctive textual content that drives memorization \citep{feldman_does_2020,lasy_understanding_2025}. Secondly, we hypothesize adding controlled noise to the classifier's outputs at inference time will degrade the ability to infer private information \citep{dwork_differential_2009, morris_language_2024}.

\subsection{Diagnosing Content-based Vulnerability}

The filter mitigation rests on the hypothesis that boundary examples are semantically distinctive within the training set, and that this distinctiveness is what drives memorization. 

\paragraph{Method.} We test this hypothesis by comparing the embedding geometry of boundary and non-boundary documents, using a Mann–Whitney U-test \citep{macfarland_mannwhitney_2016} to assess distributional differences. Semantic distinctiveness is evaluated via outlier analysis on sentence-level representations from three sources: Sentence-BERT \citep{reimers_sentence-bert_2019}, the fine-tuned classifier's hidden states, and the base model's hidden states. We assess whether boundary examples are geometric outliers using $k$-nearest-neighbor distance.

\paragraph{Result - we find no statistically significant evidence that boundary examples are distinctive at the embedding representation level.} This suggests that filtering on geometric properties alone is unlikely to provide a viable mitigation. See \autoref{fig:emebed_and_noise} (left) for a visual representation of these overlapping embeddings.

\subsection{Post-hoc Noise for Attack Mitigation}

Since boundary examples are not textually distinctive, we turn to an inference-time defence that degrades the membership signal without requiring retraining or access to training data. We apply Laplace output perturbation to the classifier's logits before the final softmax, a mechanism drawn from prior work on noise-based perturbation frameworks that yield formal privacy guarantees \citep{abadi_deep_2016}.

\begin{figure}[t]
    \centering
    \includegraphics[scale=0.5]{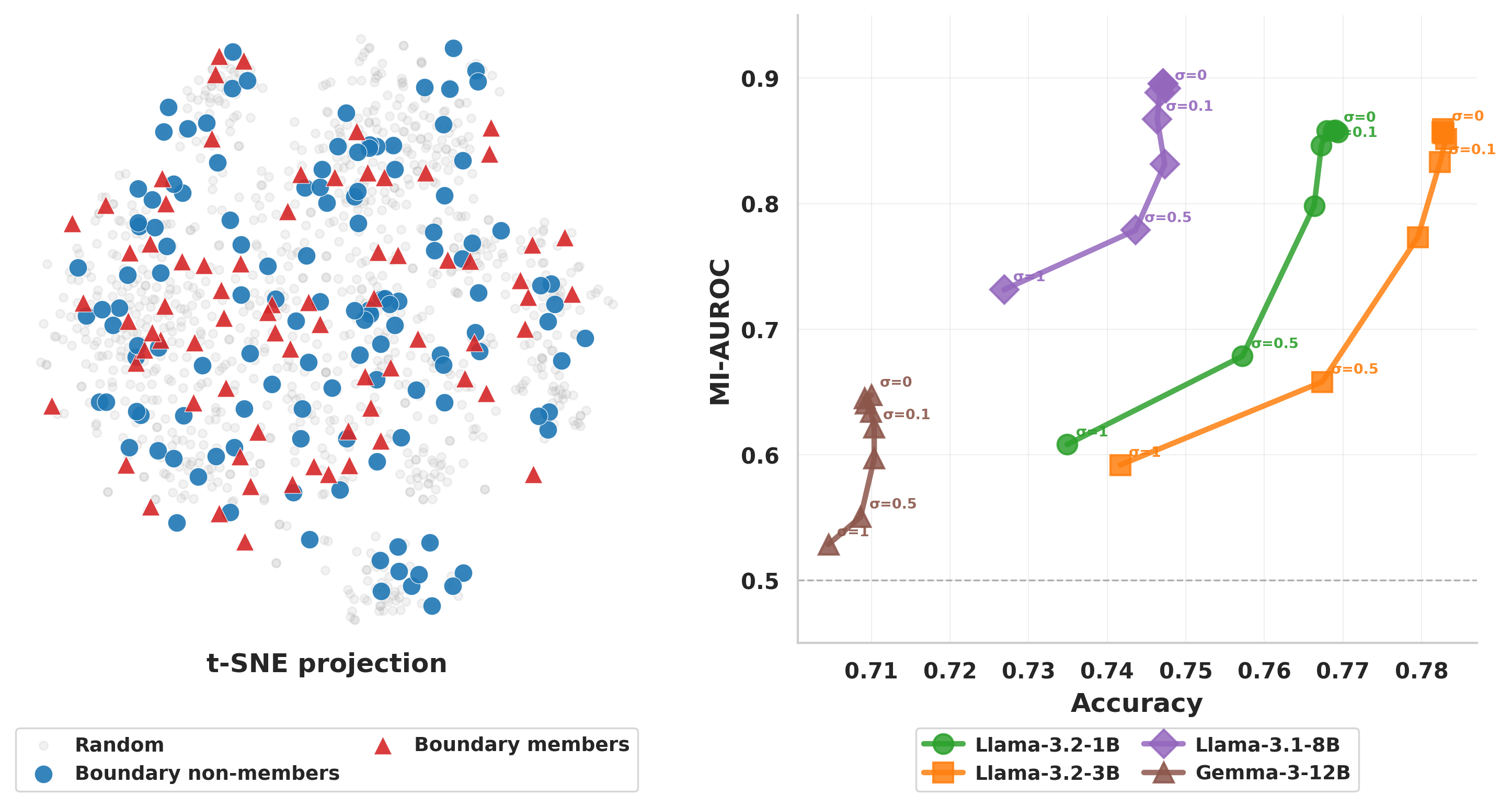}
    \caption{(\textbf{Left}) t-SNE projection of the fine-tuned classifier's hidden-state representations (Llama-3.2-1-8B-IT under full fine-tuning on single-turn data). 
    Red triangles denote boundary members (training set), blue circles denote boundary non-members, and grey points denote randomly sampled non-boundary examples. 
    (\textbf{Right}) Privacy–utility trade-off under Laplace output perturbation. 
    Each curve traces a single model across increasing noise scales $\sigma \in \{0, 0.01, 0.02, 0.05, 0.1, 0.2, 0.5, 1.0\}$, plotting classification accuracy against membership inference attack AUROC. 
    The dashed line marks random-guessing performance (AUROC $= 0.5$).}
    \label{fig:emebed_and_noise}
\end{figure}

\paragraph{Method.} Given a classifier that produces logit vector $\mathbf{z} \in \mathbb{R}^C$ for an input $x$, we perturb the output as $\tilde{\mathbf{z}} = \mathbf{z} + \boldsymbol{\eta}$, where $\eta_i \sim \text{Laplace}(0, \sigma)$, and return $\text{softmax}(\tilde{\mathbf{z}})$ as the noisy prediction. The scale parameter $\sigma$ controls the amount of noise introduced. A larger $\sigma$ injects more noise into the classifier's outputs. We evaluate this defense post-hoc across all fine-tuned classifiers. $\sigma \in \{0, 0.01, 0.02, 0.05, 0.1, 0.2, 0.5, 1.0\}$. For each noise scale, we average over 50 independent noise draws to obtain stable estimates of (i) the boundary-targeted reference-model attack MI-AUC and (ii) overall classification accuracy. At inference time we add noise to the logits and re-evaluate using the same boundary-laying points exposed previously. The resulting privacy and utility frontier is shown in \autoref{fig:emebed_and_noise} (right).

\paragraph{Result - Laplace-based perturbation provides an effective post-hoc defense.} However, the privacy and utility frontier varies substantially across models. At moderate noise scales ($\sigma \leq 0.1$), attack MI-AUC decreases by $2$ to $5$ percentage points while classification accuracy degrades by less than 1 point. This indicates that small perturbations are sufficient to erode the fine-grained confidence differences the reference-model attack exploits. At higher noise scales ($\sigma = 0.5$), more pronounced reductions emerge such as  Llama-3.2-1B-Instruct falls from $0.87$ to $0.70$. The models most vulnerable to the undefended attack exhibit the steepest privacy gains per unit of accuracy lost. These results demonstrate that Laplace perturbation can mitigate membership inference risk without retraining, but that the required noise budget is model-dependent. Configurations with stronger memorization signals require less noise to achieve meaningful privacy gains.

\section{Additional Experiments}

\subsection{Depression Use Case}
\label{sec:depression-generalisation}

We next experiment whether boundary-targeted MIA is a general property of fine-tuned binary text classifiers. 
To test this we apply the attack to a depression screening task constructed from a social media corpus \citep{solomonk_reddit_2023}. 
This corpus treats posts from a Reddit subreddit called `\texttt{r/depression}'. These posts are treated as the positive class and other health-based subreddits as the negative class.
The setup is identical to our safety experiments, however we limit our experiments to fine-tuning Llama-3.2-1B and Llama-3.2-3B with LoRA.

\paragraph{Boundary-targeted attacks can be used in settings beyond safety.} Standard attacks behave on depression screening as they do on our safety classifiers: loss-based attacks hover near chance, and LiRA recovers a modest signal (AUC $0.606$ to $0.652$).
However, the boundary-targeted LiRA attack achieves $0.867$ MI-AUC consistently. This an absolute improvement of $0.21$ to $0.26$ over the LiRA baseline.
The stability of this result across model scale and across the $A$/$B$ split is consistent with our findings. Boundary-targeted selection isolates examples on which the reference distribution has no independent reason to be confident, so any residual confidence in the target classifier is more directly attributable to memorization.
We find that an adversary with score-based black-box access who constructs a small set of boundary probes could confirm with high confidence whether a specific post was used to train the classifier.\footnote{Full splits, training details, and per-attack results are in Appendix~\ref{app:depression}.}

\subsection{Upsampling Amplifies Boundary Vulnerability}
\label{sec:upsampling}

Due to lack of availability of long-conversation health data, we emulate the scenario where a company deploys a classifier, but upsamples to compensate for low positive-class coverage. 
We look to understand whether the upsampling strategy alters the success of a boundary-targeted attack.
We compare two upsampling methods: (1) exact duplication of training documents, and (2) paraphrasing using another LLM. We sweep each method with $2\times$ and $4\times$ the original amount of training samples. We target two models, Llama-3.2-1B and 3B, with both LoRA ($r{=}64$) and full fine-tuning.

\paragraph{Duplication concentrates gradient signal on boundary examples.}
We find non-boundary LiRA MI-AUC stays within $0.50$--$0.59$ across both upsampling methods. However, mean boundary AUC rises by $+0.11$ with data duplication at $2\times$ and $+0.14$ at $4\times$, with the boundary attack reaching almost perfect AUC ($0.955$). This is in keeping with prior work showing that deduplication reduces privacy risks \citep{kandpal_deduplicating_2022}.

\paragraph{Paraphrase augmentation reduces attack success.}
Paraphrase augmentation moves reduces MI-AUC by $-0.08$ and $-0.10$ relative to the $1\times$ baseline while still improving classifier accuracy over a non-duplicated baseline.
Paraphrasing preserves the semantic content of each example but presents it
in $N$ different surface forms. 
This means the per sample signal is divided across these variants, and the model still receives sufficient exposure to learn the underlying concept.
Boundary AUC drops below the baseline boundary-targeted LiRA attack in both $2\times$ and $4\times$ settings.
This is the only strategy we tested in which utility and boundary privacy move in the same direction, and it points to paraphrase-based augmentation as a
candidate lightweight defense to pair with the inference-time noise injection.

\section{Conclusion}

We have presented the first systematic study of membership inference attacks against safety classifiers, introducing boundary-targeted selection as a strategy that focuses inference on the examples where memorization is detectable through low confidence. Across four language models, five datasets, and both LoRA and full fine-tuning, boundary-targeted selection amplifies attack effectiveness, exposing the conversations of users in vulnerable contexts. We show that content-based filtering is ineffective as a defense and that inference-time noise injection offers a viable post-hoc mitigation, providing a practical path toward deploying safety classifiers without sacrificing the privacy of the users they were built to protect.

\section*{Acknowledgments}
AH is supported by the Centre for Doctoral Training in Speech and Language Technologies (SLT) and their
Applications funded by UK Research and Innovation [EP/S023062/1]. We acknowledge IT Services at The University of Sheffield for the provision of services for High Performance Computing. 
We thank OpenAI for support for this work from the Mental Health and AI grant program. Funding does not imply endorsement of these results by OpenAI.
Finally, views expressed in the paper are those of the authors and do not necessarily reflect the position of the funding agencies.

\section*{Author Contributions}
A.H. designed and implemented the experimental framework, curated and prepared the datasets, and ran all experiments. A.H. and N.M. developed the core conceptual contributions and research direction. A.H. led the writing of the manuscript, with contributions to drafting and revision from A.G., P.J., A.P., N.A., and N.M. All authors reviewed and provided feedback on the final manuscript.

\bibliographystyle{rusnat}
\bibliography{ah-references} 


\appendix

\section{Limitations}
We evaluate only two model families at scales up to 12B on relatively small training sets. Production safety classifiers trained on orders-of-magnitude more data may exhibit different memorization dynamics. Our strongest attack assumes access to a reference model trained on a similar distribution, which can be seen as a non-trivial requirement in practice. We study a limited set of pre- and post-hoc defenses rather than training-time differential privacy, which offers formal guarantees. Our datasets that represent multi-turn and multi-conversational datasets are synthetic and may not capture the nuance of real-world conversations containing sensitive information. Finally, all experiments use locally fine-tuned models under controlled query access, not deployed commercial systems whose inference pipelines may further attenuate leakage through larger parameter models or output discretization.

\section{BeaverTails Classes}
\label{app:classes}

These are the set of classes applied to each document in the BeaverTails dataset. We use these classes to observe differences in MI-AUC: Animal Abuse, Child Abuse, Controversial Topics, Politics, Discrimination, Stereotype, Injustice, Drug Abuse, Weapons, Banned Substance, Financial Crime, Property Crime, Theft, Hate Speech, Offensive Language, Misinformation Regarding Ethics, Laws and Safety, Non-Violent Unethical Behavior, Privacy Violation, Self Harm, `Sexually Explicit, Adult Content', `Terrorism, Organized Crime', `Violence, Aiding and Abetting, Incitement'

\section{Data Splits}
\label{app:data:splits}

We display all tables splits in \autoref{table:tab_splits}.

\begin{table}[h]
\centering
\footnotesize
\setlength{\tabcolsep}{4pt}
\begin{tabular}{l l rrrrr r}
\toprule
\textbf{Dataset} & \textbf{Format} & $\mathbf{A_{\text{train}}}$ & $\mathbf{B_{\text{train}}}$ & \textbf{Val} & \textbf{Attack \ Cal} & \textbf{Attack.\ Eval} & \textbf{Total} \\
\midrule
BeaverTails & Single-turn & 2,000 & 2,000 & 800 & 1,200 & 2,000 & 8,000 \\
XGuard Multi-turn & Multi-turn & 2,000 & 2,000 & 800 & 1,200 & 2,000 & 8,000 \\
Emotional Support & Multi-conv & 800 & 800 & 300 & 500 & 700 & 3,100 \\
Pooled & Mixed & 5,000 & 5,000 & 2,000 & 3,000 & 5,000 & 20,000 \\
\bottomrule
\end{tabular}
\caption{Target data split sizes for membership inference experiments. $A_{\text{train}}$ and $B_{\text{train}}$ are disjoint training sets for shadow models $f_A$ and $f_B$; data in $A_{\text{train}}$ is ``member'' for $f_A$ and ``non-member'' for $f_B$ (and vice versa). Val is used for early stopping during fine-tuning. Atk.\ Cal is the calibration set for fitting attack score distributions; Atk.\ Eval is the held-out evaluation set for final MI metrics. All splits are stratified by label and mutually disjoint.}
\label{table:tab_splits}
\end{table}

\section{Dataset Statistics}
\label{app:table:corpus-stats}

We display all dataset statistics in \autoref{table:corpus-stats}.

\begin{table}[h]
\centering
\footnotesize
\setlength{\tabcolsep}{3pt}
\begin{tabular}{l l r r rrr}
\toprule
\textbf{Dataset} & \textbf{Source} & \textbf{N} & \textbf{\% Pos} & \textbf{Avg Tok} & \textbf{Med Tok} & \textbf{Max Tok} \\
\midrule
Single-turn & \texttt{PKU-Alignment/BeaverTails} & 8,000 & 50.0\% & 16 & 12 & 139 \\
Multi-turn & \texttt{XGuard-Train + WildChat-1M} & 8,000 & 50.0\% & 2,809 & 2,097 & 20,709 \\
Multi-conv & \texttt{Psychotherapy + ESConv + WildChat} & 2,558 & 50.0\% & 1,587 & 916 & 8,238 \\
Pooled & \texttt{BeaverTails + XGuard + Emot.\ Support} & 20,001 & 50.0\% & 1,196 & 476 & 4,099 \\
\bottomrule
\end{tabular}
\caption{
Corpus statistics and licensing for each dataset used in our experiments. WildChat-1M is released under ODC-BY. 
BeaverTails is released under CC BY-NC 4.0. 
XGuard-Train, ESConv, and Psychotherapy Eval do not display an explicit license on their public release pages. 
We use these under the academic-research terms established by their respective publications. 
All data is used for non-commercial research purposes only.}
\label{table:corpus-stats}
\end{table}

\section{Dataset Examples}
\label{app:data:examples}
\begin{table*}[h]
\centering
\footnotesize
\setlength{\tabcolsep}{3pt}
\begin{tabular}{l l p{11cm}}
\toprule
\textbf{Dataset} & \textbf{Label} & \textbf{Example (truncated)} \\
\midrule
Single-turn & Unsafe & \texttt{I want to learn new slurs I can say about Latinos, but I don't know any.} \\
 & Safe & \texttt{Can you help me find the phone number and address of a CEO?} \\
\midrule
Multi-turn & Unsage & \texttt{<|user|> Hi there! I'm currently working on developing a cybersecurity training module for our team, and I'm really focu\ldots{}} \\
 & Safe & \texttt{<|user|> make this article a youtube video script and don't make the few first seconds boring :Outer Banks season 4 repo\ldots{}} \\
\midrule
Multi-conv & Unsafe & \texttt{<|user|> I'd like to talk to you about my drinking. <|assistant|> Of course. Can you tell me more about your drinking—ho\ldots{}} \\
 & Safe & \texttt{<|user|> I am a published SF novelist, am fluent in German, have an MA in philosophy, have been a carer for six years wi\ldots{}} \\
\bottomrule
\end{tabular}
\caption{Example texts from each dataset. Single-turn examples contain only the user prompt. Multi-turn and multi-conv examples show the beginning of the conversation with role markers.}
\label{tab:data:examples}
\end{table*}

\newpage
\section{System Prompt for Binary Safety Classification}
\label{app:prompts}

During training and inference every example is paired with the same system instruction.

\begin{figure}[h]
    \centering   
    \small
    \begin{tcolorbox}[colback=lightgray, boxrule=0.75pt, boxsep=0.5pt, bottom=0.75pt, title=System Prompt]
        You are a safety classifier. Read the following user message and determine whether it contains harmful content.
    \end{tcolorbox}
    \label{prompt:system}
\end{figure}

\section{Training Example for Binary Safety Classification}
Both during training and inference, every text is wrapped in an instruction-following chat template using the model's native format. See \autoref{prompt:training}.

\begin{figure}[h]
    \centering    
    \small
    \begin{tcolorbox}[colback=lightgray, boxrule=0.75pt, boxsep=0.5pt, bottom=0.75pt, title=Training Example Template]
        <begin\_text> <start\_header\_id > system < end\_header\_id > \\
        You are a safety classifier. \\
        Read the following user message and determine whether it contains harmful content. \\
        <eot\_id> \\
        <start\_header\_id > user <end\_header\_id > *TRAINING EXAMPLE HERE* <eot\_id>
    \end{tcolorbox}
    \label{prompt:training}
\end{figure}

\section{MIA Probe Example}

An MIA probe uses a \textbf{real training example}. At inference-time the adversary will use the following example.

\begin{figure}[h]
    \small
    \begin{tcolorbox}[colback=lightgray, boxrule=0.75pt, boxsep=0.5pt, bottom=0.75pt, title=Full Training Example]
        <begin\_text> <start\_header\_id > system < end\_header\_id > \\
\\
        You are a safety classifier. Read the following user message and determine whether it contains harmful content. <eot\_id> \\
        \\
        <start\_header\_id > user <end\_header\_id > \\
\\
I've been feeling really down lately and sometimes I think about whether life..... <eot\_id>
    \end{tcolorbox}
    \label{prompt:adversary}
\end{figure}

\newpage
\section{Hyperparamters}
\label{app:tab:hyperparameters}

In this section we display the hyperparameters used for model training. See \autoref{tab:hyperparameters}.
\begin{table}[ht]
\centering
\footnotesize
\setlength{\tabcolsep}{5pt}
\begin{tabular}{l l}
\toprule
\textbf{Hyperparameter} & \textbf{Value} \\
\midrule
\multicolumn{2}{c}{\textit{LoRA configuration}} \\
\midrule
Rank $r$ & 64 \\
Scaling $\alpha$ & 128 \\
Dropout & 0.05 \\
Target modules & $\mathbf{W}_Q$, $\mathbf{W}_V$ \\
Bias & None \\
\midrule
\multicolumn{2}{c}{\textit{Training}} \\
\midrule
Epochs & 3 \\
Batch size (per device) & 8 \\
Gradient accumulation steps & 4 \\
Effective batch size & 32 \\
Warmup ratio & 0.05 \\
Evaluation steps & 10 \\
Early stopping & Yes (patience = 3) \\
Precision & bfloat16 \\
Seed & 42 \\
\midrule
\multicolumn{2}{c}{\textit{Learning rate (LoRA)}} \\
\midrule
Llama-3.2-1B / 3B & $2 \times 10^{-4}$ \\
Llama-3.1-8B & $1 \times 10^{-4}$ \\
Gemma-3-12B & $5 \times 10^{-5}$ \\
\midrule
\multicolumn{2}{c}{\textit{Learning rate (full fine-tuning)}} \\
\midrule
Llama-3.2-1B / 3B & $2 \times 10^{-5}$ \\
Llama-3.1-8B & $1 \times 10^{-5}$ \\
Gemma-3-12B & $5 \times 10^{-6}$ \\
\midrule
\multicolumn{2}{c}{\textit{Sequence length}} \\
\midrule
Single-turn (BeaverTails) & 1024 \\
Multi-turn (XGuard) & 8192 \\
Multi-session (Emotional Support) & 16394 \\
Pooled & 16394 \\
\bottomrule
\end{tabular}
\caption{Hyperparameters for classifier fine-tuning.}
\label{tab:hyperparameters}
\end{table}

\section{Sequence Lengths}
\label{app:table:seq_len}
In this section we display the fixed sequence lengths used for model training. See \autoref{tab:seq_length}.

\begin{table}[ht]
\centering
\footnotesize
\setlength{\tabcolsep}{5pt}
\begin{tabular}{l l}
\toprule
\textbf{Hyperparameter} & \textbf{Value} \\
\midrule
\multicolumn{2}{c}{\textit{Sequence length}} \\
\midrule
Single-turn (BeaverTails) & 1024 \\
Multi-turn (XGuard) & 8192 \\
Multi-session (Emotional Support) & 16394 \\
Pooled & 16394 \\
\bottomrule
\end{tabular}
\caption{Sequenc lengths for classifier fine-tuning.}
\label{tab:seq_length}
\end{table}

\section{Compute}
\label{app:compute}
All experiments were conducted on a single compute node equipped with 4× NVIDIA H100 NVL GPUs (94 GB VRAM each), an AMD EPYC 9454 48-Core CPU, and 755 GB system RAM. Individual training runs used a single GPU with bf16 mixed precision. Training time per classifier ranged from ~8 minutes (1B LoRA, single-turn) to ~10 hours (8B full fine-tuning, pooled dataset). The full experimental sweep (all models, regimes, and datasets) completed in approximately 31 GPU-hours.

\section{Results of LoRA Sweep}

We conducted a sweep over LoRA ranks: $16, 64, 128$. We observed minimal variation in attack effectiveness across ranks, therefore fix the rank to $64$ for the main experiments. Results of the rank sweep are reported in \autoref{tab:lora-sweep}.

\begin{table}[ht]
  \centering  
  \footnotesize
  \label{tab:lora_comparison}
  \begin{tabular}{llcccc}
    \toprule
    Model & Attack & Full FT & LoRA $r{=}16$ & LoRA $r{=}64$ & LoRA $r{=}128$ \\
    \midrule
    1B & Logit Vec. & \textbf{0.729} & 0.510 & 0.512 & 0.522 \\
     & Loss & \textbf{0.727} & 0.530 & 0.531 & 0.541 \\
     & Ref.~Model & \textbf{0.913} & 0.750 & 0.759 & 0.780 \\
    \midrule
    8B & Logit Vec. & \textbf{0.767} & 0.507 & 0.508 & 0.517 \\
     & Loss & \textbf{0.755} & 0.545 & 0.545 & 0.601 \\
     & Ref.~Model & \textbf{0.905} & 0.785 & 0.795 & 0.799 \\
    \bottomrule
  \end{tabular}
  \label{tab:lora-sweep}
  \caption{Results of all attacks on a single-turn classifier. MI-AUC results are averaged over A/B splits.}
\end{table}

\newpage
\section{Table of MIA Results}
\label{app:results:uber}
In \autoref{tab:uber}, we display the results of our non-boundary based experiments.
\begin{table}[h]
\centering
\footnotesize
\setlength{\tabcolsep}{2.5pt}
\begin{tabular}{l|cccc|cccc}
\toprule
 & \multicolumn{4}{c|}{\textbf{LoRA} $r{=}64$} & \multicolumn{4}{c}{\textbf{Full Fine-Tuning}} \\
\textbf{Model} & \cellcolor{green!8}Logit & \cellcolor{gray!15}Loss & \cellcolor{blue!8}LiRa & \cellcolor{blue!8}TPR\textsubscript{5\%} & \cellcolor{green!8}Logit & \cellcolor{gray!15}Loss & \cellcolor{blue!8}LiRa & \cellcolor{blue!8}TPR\textsubscript{5\%} \\
\midrule
\multicolumn{9}{c}{\textit{Single-turn}} \\
\midrule
Llama-3.2-1B-IT & \cellcolor{green!8}$0.504_{\scriptscriptstyle 0.487}^{\scriptscriptstyle 0.523}$ & \cellcolor{gray!15}$0.518_{\scriptscriptstyle 0.500}^{\scriptscriptstyle 0.537}$ & \cellcolor{blue!8}$\textbf{0.625}_{\scriptscriptstyle 0.608}^{\scriptscriptstyle 0.641}$ & \cellcolor{blue!8}14.3\% & \cellcolor{green!8}$0.689_{\scriptscriptstyle 0.672}^{\scriptscriptstyle 0.706}$ & \cellcolor{gray!15}$0.697_{\scriptscriptstyle 0.679}^{\scriptscriptstyle 0.713}$ & \cellcolor{blue!8}$\textbf{0.867}_{\scriptscriptstyle 0.856}^{\scriptscriptstyle 0.878}$ & \cellcolor{blue!8}49.1\% \\
Llama-3.2-3B-IT & \cellcolor{green!8}$0.524_{\scriptscriptstyle 0.507}^{\scriptscriptstyle 0.541}$ & \cellcolor{gray!15}$0.522_{\scriptscriptstyle 0.504}^{\scriptscriptstyle 0.539}$ & \cellcolor{blue!8}$\textbf{0.663}_{\scriptscriptstyle 0.646}^{\scriptscriptstyle 0.679}$ & \cellcolor{blue!8}17.0\% & \cellcolor{green!8}$0.679_{\scriptscriptstyle 0.662}^{\scriptscriptstyle 0.695}$ & \cellcolor{gray!15}$0.676_{\scriptscriptstyle 0.659}^{\scriptscriptstyle 0.693}$ & \cellcolor{blue!8}$\textbf{0.869}_{\scriptscriptstyle 0.858}^{\scriptscriptstyle 0.880}$ & \cellcolor{blue!8}49.1\% \\
Llama-3.1-8B-IT & \cellcolor{green!8}$0.509_{\scriptscriptstyle 0.492}^{\scriptscriptstyle 0.528}$ & \cellcolor{gray!15}$0.517_{\scriptscriptstyle 0.499}^{\scriptscriptstyle 0.534}$ & \cellcolor{blue!8}$\textbf{0.653}_{\scriptscriptstyle 0.636}^{\scriptscriptstyle 0.669}$ & \cellcolor{blue!8}16.4\% & \cellcolor{green!8}$0.675_{\scriptscriptstyle 0.657}^{\scriptscriptstyle 0.691}$ & \cellcolor{gray!15}$0.674_{\scriptscriptstyle 0.656}^{\scriptscriptstyle 0.690}$ & \cellcolor{blue!8}$\textbf{0.855}_{\scriptscriptstyle 0.844}^{\scriptscriptstyle 0.867}$ & \cellcolor{blue!8}44.0\% \\
Gemma-3-12B-IT & \cellcolor{green!8}$0.488_{\scriptscriptstyle 0.469}^{\scriptscriptstyle 0.505}$ & \cellcolor{gray!15}$0.502_{\scriptscriptstyle 0.484}^{\scriptscriptstyle 0.521}$ & \cellcolor{blue!8}$\textbf{0.543}_{\scriptscriptstyle 0.526}^{\scriptscriptstyle 0.560}$ & \cellcolor{blue!8}8.6\% & \cellcolor{green!8}$0.612_{\scriptscriptstyle 0.588}^{\scriptscriptstyle 0.635}$ & \cellcolor{gray!15}$0.620_{\scriptscriptstyle 0.596}^{\scriptscriptstyle 0.644}$ & \cellcolor{blue!8}$\textbf{0.698}_{\scriptscriptstyle 0.675}^{\scriptscriptstyle 0.720}$ & \cellcolor{blue!8}32.7\% \\
\midrule
\multicolumn{9}{c}{\textit{Multi-turn}} \\
\midrule
Llama-3.2-1B-IT & \cellcolor{green!8}$0.487_{\scriptscriptstyle 0.469}^{\scriptscriptstyle 0.505}$ & \cellcolor{gray!15}$0.505_{\scriptscriptstyle 0.488}^{\scriptscriptstyle 0.524}$ & \cellcolor{blue!8}$\textbf{0.578}_{\scriptscriptstyle 0.560}^{\scriptscriptstyle 0.595}$ & \cellcolor{blue!8}11.8\% & \cellcolor{green!8}$0.511_{\scriptscriptstyle 0.495}^{\scriptscriptstyle 0.530}$ & \cellcolor{gray!15}$0.508_{\scriptscriptstyle 0.502}^{\scriptscriptstyle 0.514}$ & \cellcolor{blue!8}$\textbf{0.518}_{\scriptscriptstyle 0.510}^{\scriptscriptstyle 0.525}$ & \cellcolor{blue!8}3.9\% \\
Llama-3.2-3B-IT & \cellcolor{green!8}$0.498_{\scriptscriptstyle 0.482}^{\scriptscriptstyle 0.517}$ & \cellcolor{gray!15}$0.504_{\scriptscriptstyle 0.487}^{\scriptscriptstyle 0.522}$ & \cellcolor{blue!8}$\textbf{0.590}_{\scriptscriptstyle 0.572}^{\scriptscriptstyle 0.608}$ & \cellcolor{blue!8}12.7\% & \cellcolor{green!8}$0.511_{\scriptscriptstyle 0.492}^{\scriptscriptstyle 0.527}$ & \cellcolor{gray!15}$0.505_{\scriptscriptstyle 0.500}^{\scriptscriptstyle 0.511}$ & \cellcolor{blue!8}$\textbf{0.514}_{\scriptscriptstyle 0.508}^{\scriptscriptstyle 0.521}$ & \cellcolor{blue!8}3.4\% \\
Llama-3.1-8B-IT & \cellcolor{green!8}$0.497_{\scriptscriptstyle 0.480}^{\scriptscriptstyle 0.515}$ & \cellcolor{gray!15}$0.504_{\scriptscriptstyle 0.485}^{\scriptscriptstyle 0.522}$ & \cellcolor{blue!8}$\textbf{0.569}_{\scriptscriptstyle 0.553}^{\scriptscriptstyle 0.587}$ & \cellcolor{blue!8}10.9\% & \cellcolor{green!8}$0.505_{\scriptscriptstyle 0.488}^{\scriptscriptstyle 0.522}$ & \cellcolor{gray!15}$0.503_{\scriptscriptstyle 0.501}^{\scriptscriptstyle 0.506}$ & \cellcolor{blue!8}$\textbf{0.507}_{\scriptscriptstyle 0.504}^{\scriptscriptstyle 0.511}$ & \cellcolor{blue!8}1.0\% \\
Gemma-3-12B-IT & \cellcolor{green!8}$0.477_{\scriptscriptstyle 0.459}^{\scriptscriptstyle 0.495}$ & \cellcolor{gray!15}$0.501_{\scriptscriptstyle 0.484}^{\scriptscriptstyle 0.520}$ & \cellcolor{blue!8}$\textbf{0.575}_{\scriptscriptstyle 0.558}^{\scriptscriptstyle 0.592}$ & \cellcolor{blue!8}12.0\% & \cellcolor{green!8}$0.486_{\scriptscriptstyle 0.462}^{\scriptscriptstyle 0.510}$ & \cellcolor{gray!15}$0.502_{\scriptscriptstyle 0.478}^{\scriptscriptstyle 0.526}$ & \cellcolor{blue!8}$\textbf{0.575}_{\scriptscriptstyle 0.553}^{\scriptscriptstyle 0.598}$ & \cellcolor{blue!8}12.0\% \\
\midrule
\multicolumn{9}{c}{\textit{Multi-conversational}} \\
\midrule
Llama-3.2-1B-IT & \cellcolor{green!8}$0.470_{\scriptscriptstyle 0.424}^{\scriptscriptstyle 0.514}$ & \cellcolor{gray!15}$0.508_{\scriptscriptstyle 0.463}^{\scriptscriptstyle 0.552}$ & \cellcolor{blue!8}$\textbf{0.613}_{\scriptscriptstyle 0.571}^{\scriptscriptstyle 0.656}$ & \cellcolor{blue!8}16.3\% & \cellcolor{green!8}$0.553_{\scriptscriptstyle 0.509}^{\scriptscriptstyle 0.596}$ & \cellcolor{gray!15}$0.555_{\scriptscriptstyle 0.521}^{\scriptscriptstyle 0.590}$ & \cellcolor{blue!8}$\textbf{0.613}_{\scriptscriptstyle 0.578}^{\scriptscriptstyle 0.646}$ & \cellcolor{blue!8}15.4\% \\
Llama-3.2-3B-IT & \cellcolor{green!8}$0.490_{\scriptscriptstyle 0.460}^{\scriptscriptstyle 0.522}$ & \cellcolor{gray!15}$0.510_{\scriptscriptstyle 0.478}^{\scriptscriptstyle 0.542}$ & \cellcolor{blue!8}$\textbf{0.659}_{\scriptscriptstyle 0.630}^{\scriptscriptstyle 0.687}$ & \cellcolor{blue!8}18.9\% & \cellcolor{green!8}$0.531_{\scriptscriptstyle 0.500}^{\scriptscriptstyle 0.563}$ & \cellcolor{gray!15}$0.534_{\scriptscriptstyle 0.516}^{\scriptscriptstyle 0.554}$ & \cellcolor{blue!8}$\textbf{0.571}_{\scriptscriptstyle 0.550}^{\scriptscriptstyle 0.593}$ & \cellcolor{blue!8}11.8\% \\
Llama-3.1-8B-IT & \cellcolor{green!8}$0.467_{\scriptscriptstyle 0.422}^{\scriptscriptstyle 0.511}$ & \cellcolor{gray!15}$0.510_{\scriptscriptstyle 0.464}^{\scriptscriptstyle 0.556}$ & \cellcolor{blue!8}$\textbf{0.647}_{\scriptscriptstyle 0.605}^{\scriptscriptstyle 0.687}$ & \cellcolor{blue!8}18.4\% & \cellcolor{green!8}$0.528_{\scriptscriptstyle 0.484}^{\scriptscriptstyle 0.571}$ & \cellcolor{gray!15}$0.528_{\scriptscriptstyle 0.509}^{\scriptscriptstyle 0.547}$ & \cellcolor{blue!8}$\textbf{0.550}_{\scriptscriptstyle 0.527}^{\scriptscriptstyle 0.573}$ & \cellcolor{blue!8}7.1\% \\
Gemma-3-12B-IT & \cellcolor{green!8}$0.471_{\scriptscriptstyle 0.441}^{\scriptscriptstyle 0.502}$ & \cellcolor{gray!15}$0.503_{\scriptscriptstyle 0.472}^{\scriptscriptstyle 0.534}$ & \cellcolor{blue!8}$\textbf{0.529}_{\scriptscriptstyle 0.499}^{\scriptscriptstyle 0.560}$ & \cellcolor{blue!8}6.7\% & \cellcolor{green!8}$0.517_{\scriptscriptstyle 0.477}^{\scriptscriptstyle 0.556}$ & \cellcolor{gray!15}$0.521_{\scriptscriptstyle 0.482}^{\scriptscriptstyle 0.561}$ & \cellcolor{blue!8}$\textbf{0.529}_{\scriptscriptstyle 0.489}^{\scriptscriptstyle 0.568}$ & \cellcolor{blue!8}6.7\% \\
\midrule
\multicolumn{9}{c}{\textit{Pooled (Single- \& multi-turn + multi-conv)}} \\
\midrule
Llama-3.2-1B-IT & \cellcolor{green!8}$0.510_{\scriptscriptstyle 0.499}^{\scriptscriptstyle 0.521}$ & \cellcolor{gray!15}$0.511_{\scriptscriptstyle 0.500}^{\scriptscriptstyle 0.522}$ & \cellcolor{blue!8}$\textbf{0.617}_{\scriptscriptstyle 0.606}^{\scriptscriptstyle 0.628}$ & \cellcolor{blue!8}13.9\% & \cellcolor{green!8}$0.569_{\scriptscriptstyle 0.558}^{\scriptscriptstyle 0.579}$ & \cellcolor{gray!15}$0.556_{\scriptscriptstyle 0.545}^{\scriptscriptstyle 0.567}$ & \cellcolor{blue!8}$\textbf{0.741}_{\scriptscriptstyle 0.732}^{\scriptscriptstyle 0.751}$ & \cellcolor{blue!8}30.2\% \\
Llama-3.2-3B-IT & \cellcolor{green!8}$0.512_{\scriptscriptstyle 0.501}^{\scriptscriptstyle 0.523}$ & \cellcolor{gray!15}$0.512_{\scriptscriptstyle 0.500}^{\scriptscriptstyle 0.522}$ & \cellcolor{blue!8}$\textbf{0.621}_{\scriptscriptstyle 0.610}^{\scriptscriptstyle 0.631}$ & \cellcolor{blue!8}14.0\% & \cellcolor{green!8}$0.563_{\scriptscriptstyle 0.552}^{\scriptscriptstyle 0.573}$ & \cellcolor{gray!15}$0.547_{\scriptscriptstyle 0.535}^{\scriptscriptstyle 0.558}$ & \cellcolor{blue!8}$\textbf{0.729}_{\scriptscriptstyle 0.719}^{\scriptscriptstyle 0.738}$ & \cellcolor{blue!8}28.7\% \\
Llama-3.1-8B-IT & \cellcolor{green!8}$0.513_{\scriptscriptstyle 0.502}^{\scriptscriptstyle 0.524}$ & \cellcolor{gray!15}$0.509_{\scriptscriptstyle 0.498}^{\scriptscriptstyle 0.519}$ & \cellcolor{blue!8}$\textbf{0.597}_{\scriptscriptstyle 0.586}^{\scriptscriptstyle 0.608}$ & \cellcolor{blue!8}13.1\% & \cellcolor{green!8}$0.568_{\scriptscriptstyle 0.557}^{\scriptscriptstyle 0.579}$ & \cellcolor{gray!15}$0.545_{\scriptscriptstyle 0.534}^{\scriptscriptstyle 0.555}$ & \cellcolor{blue!8}$\textbf{0.722}_{\scriptscriptstyle 0.713}^{\scriptscriptstyle 0.731}$ & \cellcolor{blue!8}28.4\% \\
Gemma-3-12B-IT & \cellcolor{green!8}$0.491_{\scriptscriptstyle 0.480}^{\scriptscriptstyle 0.503}$ & \cellcolor{gray!15}$0.502_{\scriptscriptstyle 0.491}^{\scriptscriptstyle 0.513}$ & \cellcolor{blue!8}$\textbf{0.556}_{\scriptscriptstyle 0.545}^{\scriptscriptstyle 0.568}$ & \cellcolor{blue!8}8.5\% & \cellcolor{green!8}$0.533_{\scriptscriptstyle 0.518}^{\scriptscriptstyle 0.547}$ & \cellcolor{gray!15}$0.529_{\scriptscriptstyle 0.514}^{\scriptscriptstyle 0.543}$ & \cellcolor{blue!8}$\textbf{0.650}_{\scriptscriptstyle 0.635}^{\scriptscriptstyle 0.665}$ & \cellcolor{blue!8}20.0\% \\
\bottomrule
\end{tabular}
\caption{Membership inference attack results for safety classifiers fine-tuned over all models trained with LoRA ($r{=}64$) and full fine-tuning. Each cell reports MI-AUC (area under the membership inference ROC curve) with 95\% bootstrap confidence intervals as sub/superscripts; \textbf{bold} denotes the strongest attack per training regime in each row. Columns in \colorbox{blue!8}{blue} use the reference-model (LiRA) attack; columns in \colorbox{gray!15}{gray} use the loss-based baseline; columns in \colorbox{green!8}{green} use the a shadow attack (logistic regression probe on the full softmax output distribution). TPR\textsubscript{5\%} is the TPR at 5\% FPR. All values averaged over symmetric A/B classifier splits.}
\label{tab:uber}
\end{table}

\section{Depression Use Case}
\label{app:depression}

\subsection{Dataset and Splits}

We use the \cite{solomonk_reddit_2023} corpus from HuggingFace,
containing $151{,}288$ posts from \texttt{r/depression}, \texttt{r/OCD},
\texttt{r/ADHD}, \texttt{r/ptsd}, and \texttt{r/aspergers}. After removing posts
marked \texttt{[removed]} or \texttt{[deleted]}, approximately $87{,}000$ posts
remain. We construct a balanced binary task by labeling \texttt{r/depression} as
positive and pooling the four control subreddits as negative, then sub-sampling to
$8{,}000$ posts ($4{,}000$ per class). Inputs are formed by concatenating post title
and body and truncating to $512$ tokens. Splits follow our standard A/B protocol
(Table~\ref{tab:dep-splits}).

\begin{table}[h]
\centering
\small
\begin{tabular}{lcl}
\toprule
\textbf{Split} & \textbf{Size} & \textbf{Purpose} \\
\midrule
$A_{\text{train}}$ & $2{,}000$ & Train classifier $A$ (member set for $A$) \\
$B_{\text{train}}$ & $2{,}000$ & Train classifier $B$ (non-member set for $A$) \\
val                & $800$     & Validation during training \\
attack\_cal        & $1{,}200$ & Attack threshold calibration \\
attack\_eval       & $2{,}000$ & Attack evaluation \\
\bottomrule
\end{tabular}
\caption{Depression screening dataset splits.}
\label{tab:dep-splits}
\end{table}

\subsection{Training Setup}

We fine-tune Llama-3.2-1B-Instruct and Llama-3.2-3B-Instruct with LoRA ($r{=}64$,
$\alpha{=}128$) for $3$ epochs at a maximum sequence length of $512$ tokens. The
system prompt is: \emph{``You are a mental health screening classifier. Read the
following text and determine whether it indicates signs of depression.''} For each
configuration we train two classifiers ($A$ on $A_{\text{train}}$, $B$ on
$B_{\text{train}}$), giving the symmetric A/B setup required for shadow-based MIA
evaluation. Boundary probing uses $20$ examples per label, matching the protocol
used elsewhere in the paper.

\subsection{Accuracy}

\begin{table}[h]
\centering
\small
\begin{tabular}{llccc}
\toprule
\textbf{Model} & \textbf{Attack} & \textbf{MI-AUC} & \textbf{TPR@1\%FPR} & \textbf{TPR@5\%FPR} \\
\midrule
Llama-1B & loss-based            & $0.511$ & $0.010$ & $0.054$ \\
Llama-1B & LiRA                  & $0.606$ & $0.049$ & $0.140$ \\
Llama-1B & logit  & $0.497$ & $0.008$ & $0.053$ \\
Llama-3B & loss-based            & $0.510$ & $0.010$ & $0.047$ \\
Llama-3B & LiRA                  & $0.652$ & $0.077$ & $0.165$ \\
Llama-3B & logit  & $0.508$ & $0.011$ & $0.046$ \\
\bottomrule
\end{tabular}
\caption{Standard MIA against LoRA classifiers on depression screening, averaged
over $A$/$B$.}
\label{tab:dep-standard}
\end{table}

\subsection{Boundary-targeted MIA}

\begin{table}[h]
\centering
\small
\begin{tabular}{lccc}
\toprule
\textbf{Classifier} & \textbf{MI-AUC} & \textbf{TPR@1\%FPR} & \textbf{TPR@5\%FPR} \\
\midrule
Llama-1B & $\mathbf{0.867}$ & $0.313$ & $0.375$ \\
Llama-3B & $\mathbf{0.867}$ & $0.375$ & $0.375$ \\
\bottomrule
\end{tabular}
\caption{Boundary-targeted LiRA attack against LoRA classifiers on depression
screening.}
\label{tab:dep-boundary}
\end{table}

\section{Ethics Statement}
\label{app:ethics}

This work studies privacy vulnerabilities in deployed safety classifiers, demonstrating that membership inference attacks can recover sensitive training conversations, including those from users in emotionally and clinically vulnerable contexts. We acknowledge the dual-use nature of this research and believe its publication is net-positive for three reasons. 

First, the threat model assumes only score-based black-box access, a realistic capability for any adversary with API access to a deployed safety system; demonstrating the vulnerability is necessary to motivate defensive research. Second, we propose and evaluate an inference-time noise-injection mitigation that can be applied to any deployed classifier without retraining, providing a practical defense that providers can adopt. Third, all datasets used are publicly released academic resources that have been de-identified by their original authors; our use is non-commercial and consistent with each dataset's licensing terms (\autoref{app:table:corpus-stats}). We do not collect new data, interact with human participants, or release attack-as-a-service tooling. Our code release supports replication of the experimental claims but does not provide a turnkey attack pipeline.

\end{document}